\documentclass[conference]{IEEEtran}
\usepackage{amsmath}
\usepackage{amssymb}
\usepackage{amsfonts}
\usepackage{graphicx}
\usepackage{textcomp}
\usepackage{booktabs}
\usepackage{float}
\usepackage{multirow}
\usepackage{subcaption}
\usepackage{afterpage}
\usepackage{placeins}
\usepackage{cite}

\begin{document}

\title{BehaveRec: When Does Long-Term Behavioral History Help Sequential Recommendation? A Study with Session, History, and Community Signals}

\author{
\IEEEauthorblockN{Aysenur Inan,
Reza Yousefi Maragheh,
Jayesh Uddhav Kudase, \\
Selene Xu, Kaushiki Nag, Jianpeng Xu, Sushant Kumar}
\IEEEauthorblockA{Walmart Global Tech, 
Sunnyvale, CA, USA\\
\{aysenur.inan, reza.yousefimaragheh, jayeshuddhav.kudase,\\ yue.xu, kaushiki.nag, jianpeng.xu, sushant.kumar\}@walmart.com}
}

\maketitle

\begin{abstract}
Sequential recommendation models typically predict a user's next action
from recent session interactions, which limits personalization when
short-term signals are weak or ambiguous. Rather than assuming that
incorporating longer-term history and community signals is universally
beneficial, we study \emph{when} it actually helps. We propose BehaveRec,
a three-signal framework combining a causal session transformer, a
bidirectional User History Encoder (UHE) over a 90-day window, and a
GraphSAGE module over a user-user graph built from behavioral embedding
similarity rather than item co-occurrence; a learned gate blends the
three signals per user.

On an industrial dataset of 137M sequences from 12.6M users, BehaveRec
improves R@5 by $+3.2\%$ over SASRec, with a larger-capacity SASRec
baseline (more parameters than BehaveRec) still trailing by $+2.5\%$,
ruling out capacity as the explanation. More importantly, a user-level
stratification by history length reveals a clear dose-response
relationship: BehaveRec's advantage is negligible below roughly 10
interactions and grows monotonically with history richness, reaching
$+0.017$ R@5 for users with 101+ events. This pattern is corroborated
across one industrial and three public datasets spanning history-rich
(MovieLens) to history-sparse (Amazon) regimes, where session-only
models remain preferable. These headline results, and the
resulting deployment threshold, are established for category-level
intent prediction; a preliminary, single-run check on the full
item-level vocabulary (Section~\ref{sec:item_level}) yields a
directionally mixed result and is scoped as an open question rather
than a claim. We use these results to derive a practical deployment
rule -- roughly 20 interactions per user -- for when long-term history
and behavioral-graph signals are worth their added complexity. Despite this added complexity, BehaveRec remains lightweight: it uses 5.67M parameters and trains on a single GPU in under
two days.
\end{abstract}


\section{Introduction}

Recommender systems are central to the user experience of
large-scale e-commerce platforms, where predicting a customer's next
intent translates directly to measurable business outcomes. The
dominant paradigm for sequential recommendation has converged on
self-attention-based models such as SASRec~\cite{sasrec}, which model
the causal structure of a user's recent session to identify the most
likely next action.

Despite strong aggregate performance, session-only models have a fundamental
limitation: they cannot personalize predictions across users who share
identical short-term sessions. Consider two customers who both viewed and
added a coffee grinder to cart. One is a daily coffee enthusiast with 200
events across coffee equipment and pet dental care; the other is a college
student making a one-time purchase. A session-only model assigns both users
identical predictions. The session captures what the user is doing right
now; it says nothing about who the user is.

This gap motivates the integration of longer-term personal history. User
behavior over a 90-day window encodes a behavioral persona: recurring
clusters of intent, activity patterns, and category affinities that define
a user's preferences independent of any single session. Beyond personal
history, behavioral communities offer a complementary signal -- users who
share similar long-term behavioral personas tend to follow similar
transition patterns across intent clusters.

Rather than assuming richer history and community signals are always
worth their added modeling and serving complexity, we ask a more specific
question: \emph{when} does long-term behavioral history help session-based
recommendation, and by how much? Answering this requires an architecture
capable of using such signals, controlled comparisons that isolate their
marginal contribution, and an evaluation designed to detect the point at
which they stop mattering. We build BehaveRec to serve as this instrument,
and use it to conduct a user-level and cross-domain study of
history-richness effects at industrial scale. As we show, the answer is
not ``always'' or ``never'' but a quantifiable dose-response curve
(Section~\ref{sec:history_strat}) -- and we report the cases where the
added signal does \emph{not} help (Sections~\ref{sec:cross_domain},
\ref{sec:item_level}) with the same care as the cases where it does.

\subsection{Limitations of Existing Approaches}

Session-only models~\cite{sasrec,bert4rec} operate exclusively on short
interaction windows, degrading for cold-session users and multi-category
sessions. Graph-based approaches such as LightGCN~\cite{lightgcn} model
user-item co-purchase graphs, but item co-purchase similarity is a coarse
proxy for behavioral similarity: two users who purchased the same item may
have fundamentally different behavioral personas and transition tendencies.
LLM-based approaches~\cite{tallrec} are impractical for production systems
at scales of 137 million training sequences.

\subsection{Our Approach}

We propose BehaveRec, a three-signal framework encoding:
\begin{itemize}
\item \textbf{$h_{\text{short}}$ (Session Encoder):} A 4-layer causal
  transformer over the user's 7-day session, capturing immediate intent.
\item \textbf{$h_{\text{long}}$ (User History Encoder, UHE):} A 4-layer
  bidirectional transformer with a learnable \texttt{[USER]} token,
  encoding the 90-day behavioral persona into a single 256-d vector
  that is user-specific even when sessions are identical.
\item \textbf{$h_{\text{graph}}$ (Behavioral Community Signal):} A
  GraphSAGE~\cite{graphsage} aggregation over a user-user graph whose
  edges reflect behavioral similarity of learned history embeddings --
  not item co-occurrence.
\end{itemize}
A learned ThreeWayGate blends all three signals per user. BehaveRec is
trained with label-smoothed cross-entropy augmented by InfoNCE~\cite{infonce}
contrastive alignment between session and history representations. We use
this architecture as a controlled instrument: each signal can be
independently disabled, measured, and stratified by user history length,
which is what makes the empirical study in Section~\ref{sec:history_strat}
possible.

\subsection{Contributions}

\begin{itemize}
\item \textbf{A systematic study of when long-term history and
  behavioral-graph signals help.} A user-level stratification on the
  industrial test set (Table~\ref{tab:history_bins},
  Section~\ref{sec:history_strat}) shows BehaveRec's advantage over a
  session-only baseline is negligible below $\sim$10 interactions and
  grows monotonically with history length, corroborated by a
  four-dataset cross-domain comparison (Section~\ref{sec:cross_domain})
  that yields a practical deployment threshold of $\approx$20
  interactions per user.
\item \textbf{BehaveRec architecture:} session encoding, 90-day
  personal history encoding, and behavioral community graph signal
  fused via a learned ThreeWayGate.
\item \textbf{Behavioral community graph with controlled validation:}
  users are connected via cosine similarity of learned UHE history
  embeddings rather than item co-occurrence; a controlled ablation
  (Table~\ref{tab:graph_ablation}, Section~\ref{sec:graph_ablation})
  shows this outperforms item co-occurrence edges (LightGCN-style).
\item \textbf{Industrial-scale validation with capacity controls:} a
  SASRec-Large ablation (Section~\ref{sec:main_results}) confirms most
  of BehaveRec's gain is architectural rather than a capacity artifact.
\item \textbf{An honest account of where the approach does not help:}
  sparse-history Amazon benchmarks (avg 14--17 events) favor SASRec,
  and a full item-level MovieLens evaluation
  (Section~\ref{sec:item_level}) shows a directionally mixed result
  rather than a uniform win.
\item \textbf{Practical scalability:} 5.67M parameters, single A100
  GPU, under two days -- roughly a 1,000$\times$ parameter reduction
  vs.\ LLM approaches.
\end{itemize}

\section{Related Work}

\subsection{Sequential Recommendation}
Early approaches used Markov chains and matrix factorization.
GRU4Rec~\cite{gru4rec} applied recurrent units; Caser~\cite{caser}
used convolutional filters. SASRec~\cite{sasrec} introduced causal
self-attention and remains a standard baseline. BERT4Rec~\cite{bert4rec}
extended this with bidirectional attention. A common limitation is
exclusive reliance on the current session: users sharing identical
recent sequences receive identical predictions regardless of differing
long-term histories.

\subsection{Graph-Augmented Recommendation}
LightGCN~\cite{lightgcn} propagates embeddings through bipartite user-item
graphs. GraphSAGE~\cite{graphsage}, which underlies our graph module, learns
inductive aggregation functions suited for production systems with continuous
new users. SURGE~\cite{surge} builds metric-learned item graphs from sessions
for sparse settings. All these approaches build graphs from item
co-occurrence or user-item overlap. BehaveRec instead constructs
user-user graphs from behavioral similarity of learned 90-day history
embeddings, connecting users whose full behavioral patterns are similar
rather than users who merely co-purchased an item.

\subsection{LLM-Based Recommendation}
TALLRec~\cite{tallrec} and CoLLM~\cite{collm} fine-tune or augment LLM
backbones for recommendation. These approaches are impractical at
scales of 137M training sequences and tens of millions of users.
BehaveRec achieves comparable architectural goals with 5.67M
parameters on a single GPU.

\subsection{Industrial Long-Term User Modeling and User-User Graphs}
\label{sec:novelty}
Two lines of prior work are closest to BehaveRec. Industrial long-/short-term
systems (MIMN~\cite{mimn}; SIM and TWIN~\cite{sim,twin}; SLi-Rec~\cite{slirec})
combine long- and short-term interest via attention or memory networks, but
operate over raw item vocabularies through auxiliary retrieval over a sparse
subset of historical items; the resulting long-term signal is a re-weighted
item history, not a separately learnable user representation. BehaveRec
instead encodes the full 90-day window into a single bidirectional embedding
over a pre-clustered intent vocabulary and reuses that embedding for both
fusion and graph construction. Separately, embedding-based user-user
collaborative filtering~\cite{social_rec_survey, padate2017hybrid} links users by latent-factor
or social-tie similarity derived from static rating or co-purchase matrices,
so users are connected only if they overlap on items. BehaveRec's edges
instead come from cosine similarity between \emph{learned sequential history
embeddings} produced by the same UHE that drives fusion, so two users with no
items in common can still be neighbors if their 90-day transition patterns
are similar, and the graph is naturally recomputed as those embeddings drift.
We view this combination -- one shared history encoder feeding both a fusion
signal and an inductive, retraining-free behavioral graph -- as
\emph{architecturally distinct} from these two lines of prior work, rather
than a straightforward instance of either. We stop short of a claim of
established superiority: as noted in Section~\ref{sec:baselines}, we do not
benchmark directly against these systems, since they operate at ad-retrieval
scale and are not directly portable to our intent-cluster task, so we have no
head-to-head number to support a stronger claim here.

\subsection{User History Modeling}
We adopt a bidirectional transformer with a \texttt{[USER]} aggregation
token for the UHE, analogous to BERT's \texttt{[CLS]}, and InfoNCE~\cite{infonce}
contrastive alignment to pull session and history embeddings together for
the same user.

\section{Problem Formulation}

\subsection{Intent Clusters}
\label{sec:intent_clusters}
Our production setting operates over $|\mathcal{I}| = 2{,}371$ intent
clusters constructed via a three-stage LLM-assisted pipeline: (1) LLM
generates primary intent statements per product; (2) $k$-means clusters
similar intents; (3) LLM generates human-readable cluster names (e.g.,
\textit{Coffee Machine Essentials}). This reduces vocabulary size and
produces interpretable, actionable predictions. The clustering pipeline
itself is not a contribution of this work.

\subsection{Notation}
Each interaction is a pair $(c, a)$ where $c \in \mathcal{I}$ is an intent
cluster and $a \in \{\textsc{view}, \textsc{atc}, \textsc{trx}\}$ is the
activity type. For user $u$:
\begin{align}
S_u &= [(c_1,a_1),\ldots,(c_n,a_n)], \quad n \le 20 \quad \text{(7-day session)}\\
H_u &= [(c_1,a_1),\ldots,(c_m,a_m)], \quad m \le 100 \quad \text{(90-day history)}
\end{align}

\subsection{Behavioral Neighborhood Graph}
We define $G = (\mathcal{U}, E)$ with edge weight:
\begin{equation}
w(u, v) = \frac{\mathbf{h}^{long}_u \cdot \mathbf{h}^{long}_v}{\lVert \mathbf{h}^{long}_u \rVert \, \lVert \mathbf{h}^{long}_v \rVert}
\end{equation}
For each user we define $\mathcal{N}(u)$ as the top-$K=20$ most behaviorally
similar users. Unlike LightGCN~\cite{lightgcn}, two users may have no items
in common and still receive high edge weight if their 90-day behavioral
patterns are similar.

\subsection{Task Definition}
Given $S_u$, $H_u$, and $\mathcal{N}(u)$, predict:
\begin{equation}
P\left(i_t \mid S_u, H_u, \mathcal{N}(u); \Theta\right)
\end{equation}

\section{BehaveRec Architecture}

\FloatBarrier
\afterpage{%
  \begin{figure*}[!t]
    \centering
    \includegraphics[width=0.94\linewidth]{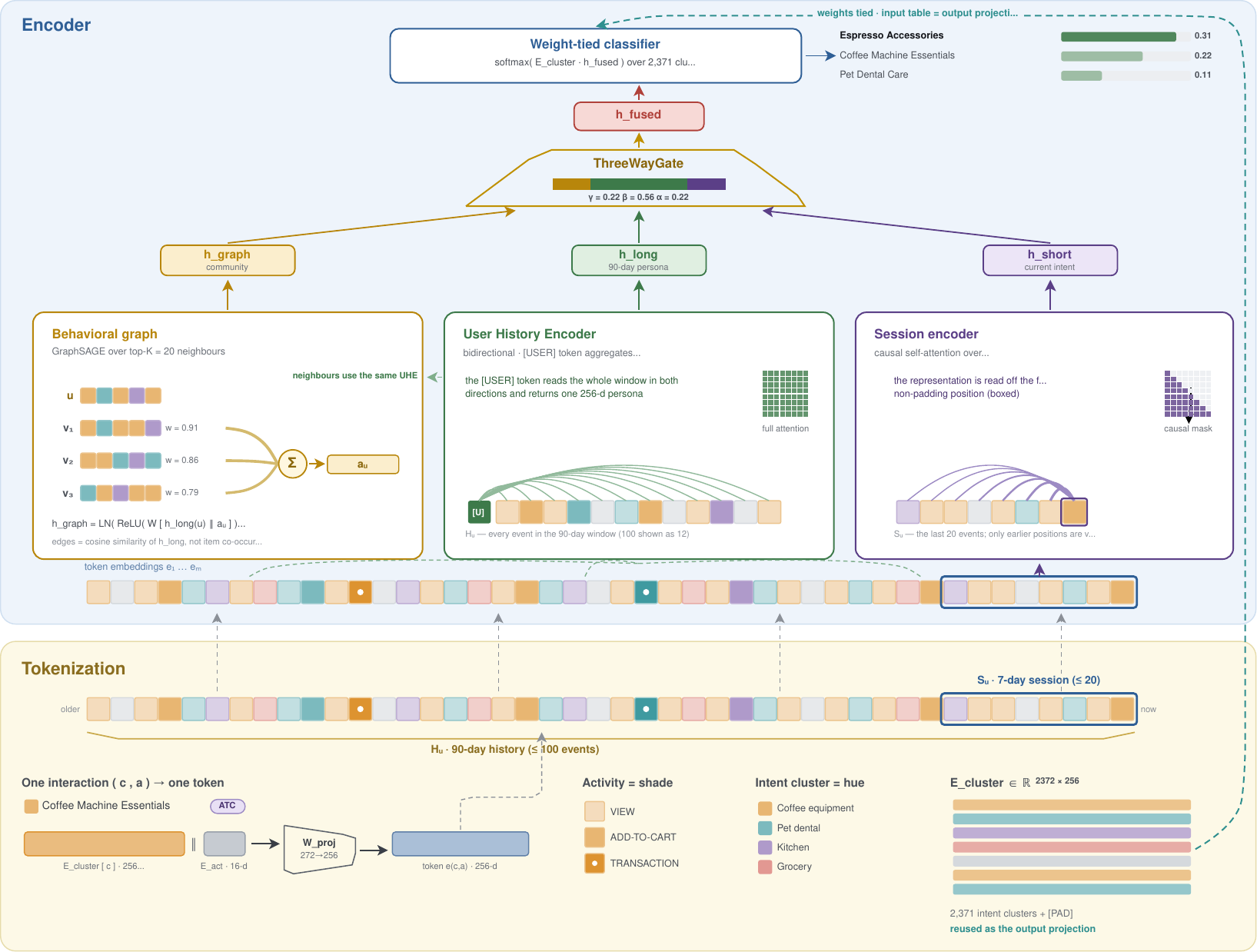}
    \caption{BehaveRec architecture. Three signals -- session $S_u$,
             history $H_u$, and neighbors $N(u)$ -- are encoded and
             fused via ThreeWayGate into $\mathbf{h}_{\text{fused}}$,
             classified over 2,371 intent clusters via a weight-tied
             classifier. UHE weights are shared between history encoding
             and GraphSAGE neighbor encoding.}
    \label{fig:architecture}
  \end{figure*}%
}

BehaveRec encodes three complementary signals and fuses them via a
learned gate. Figure~\ref{fig:architecture} illustrates the full data
flow.

BehaveRec encodes three complementary signals and fuses them via a
learned gate. Figure~\ref{fig:architecture} illustrates the full data
flow. All encoders share a common cluster and activity embedding table.

\subsection{Shared Embeddings}
\begin{equation}
\mathbf{e}_{(c,a)} = W_{\text{proj}} \left[ \mathbf{E}_{\text{cluster}}[c] \,\|\, \mathbf{E}_{\text{act}}[a] \right]
\end{equation}
where $\mathbf{E}_{\text{cluster}} \in \mathbb{R}^{(|\mathcal{I}|+1) \times 256}$
and $\mathbf{E}_{\text{act}} \in \mathbb{R}^{4 \times 16}$. The cluster
embedding table is shared with the output classifier
(Section~\ref{sec:weighttied}).

\subsection{Session Encoder ($h_{\text{short}}$)}
A 4-layer causal Transformer (4 heads, FFN=512, $d=256$), analogous to
SASRec~\cite{sasrec} but with greater capacity:
\begin{align}
\tilde{\mathbf{X}} &= \text{LayerNorm}(\mathbf{X}) \\
\mathbf{X}' &= \mathbf{X} + \text{MultiHead}(\tilde{\mathbf{X}}, \tilde{\mathbf{X}}, \tilde{\mathbf{X}}; \text{causal}) \\
\mathbf{X}'' &= \mathbf{X}' + \text{FFN}(\text{LayerNorm}(\mathbf{X}'))
\end{align}
The representation at the last non-padding position is extracted and
layer-normalized to produce $h_{\text{short}} \in \mathbb{R}^{256}$.

\subsection{User History Encoder ($h_{\text{long}}$)}
A 4-layer \emph{bidirectional} Transformer with a prepended learnable
\texttt{[USER]} token:
\begin{equation}
\mathbf{X}_{\text{uhe}} = \left[ \mathbf{e}_{\texttt{[USER]}} \,;\, \mathbf{e}_{(c_1,a_1)}, \ldots, \mathbf{e}_{(c_m,a_m)} \right]
\end{equation}
\begin{equation}
h_{\text{long}}(u) = \text{LayerNorm}\Big( \text{UHE}(\mathbf{X}_{\text{uhe}})\big[0\big] \Big)
\end{equation}
The \texttt{[USER]} token attends to every event in the history window,
aggregating recurring patterns. The same UHE (shared weights) encodes
neighbor histories, ensuring all representations live in a consistent
256-d space.

\subsection{Behavioral Graph Construction}
We L2-normalize all $h_{\text{long}}$ embeddings and build an IVF
approximate nearest-neighbor index, retrieving top-$(K{+}1)$ neighbors
per user (discarding self-matches, $K=20$). This completes in under 40
minutes on a single GPU for the full user population, and produces a
graph that is recomputed periodically as $h_{\text{long}}$ embeddings
drift with new user behavior.

\subsection{GraphSAGE Aggregation ($h_{\text{graph}}$)}
\begin{equation}
\mathbf{a}_u = \sum_{v \in \mathcal{N}(u)} \frac{w(u,v)}{\sum_{v'} w(u,v')} \; h_{\text{long}}(v)
\end{equation}
\begin{equation}
h_{\text{graph}}(u) = \text{LayerNorm}\Big( \text{Dropout}\big( \text{ReLU}\big( W_{\text{sage}} \left[ h_{\text{long}}(u) \,\|\, \mathbf{a}_u \right] \big) \big) \Big)
\end{equation}
When $\mathcal{N}(u) = \emptyset$, $\mathbf{a}_u = \mathbf{0}$ and the
gate automatically down-weights $h_{\text{graph}}$.

\subsection{ThreeWayGate}
\begin{align}
[\alpha, \beta, \gamma] &= \text{softmax}\Big( W_{\text{gate}} \left[ h_{\text{short}} \,\|\, h_{\text{long}} \,\|\, h_{\text{graph}} \right] \Big) \\
h_{\text{fused}} &= \text{LayerNorm}\Big( \alpha h_{\text{short}} + \beta h_{\text{long}} + \gamma h_{\text{graph}} \Big)
\end{align}
The gate learns distinct fusion strategies: long sessions upweight $\alpha$;
cold sessions with rich history upweight $\beta$; tight behavioral
communities upweight $\gamma$.

\subsection{Weight-Tied Classifier}
\label{sec:weighttied}
\begin{equation}
P(i_t = c \mid S_u, H_u, \mathcal{N}(u)) = \text{softmax}\Big( \mathbf{E}_{\text{cluster}}[:|\mathcal{I}|] \cdot h_{\text{fused}} \Big)_c
\end{equation}
Weight tying halves parameters attributable to the cluster vocabulary and
enforces consistency between input and output cluster representations.

\subsection{Training Objective}
\textbf{Label-smoothed cross-entropy} ($\epsilon=0.1$):
\begin{equation}
\mathcal{L}_{\text{CE}} = -(1-\epsilon) \log p_y - \frac{\epsilon}{|\mathcal{I}|} \sum_{c} \log p_c
\end{equation}

\textbf{InfoNCE alignment} (temperature $\tau=0.5$, batch size $B$):
\begin{equation}
\mathcal{L}_{\text{NCE}} = -\frac{1}{B}\sum_{i=1}^{B} \log \frac{\exp(\hat{h}^{i}_{\text{short}} \cdot \hat{h}^{i}_{\text{long}} / \tau)}{\sum_{j=1}^{B} \exp(\hat{h}^{i}_{\text{short}} \cdot \hat{h}^{j}_{\text{long}} / \tau)}
\end{equation}

\textbf{Final:} $\mathcal{L} = \mathcal{L}_{\text{CE}} + 0.1\,\mathcal{L}_{\text{NCE}}$

\subsection{Implementation Details}
\label{sec:impl}
$d=256$, 4 Transformer layers, 4 heads, FFN=512, activity embedding 16-d,
5.67M parameters. AdamW with lr $3{\times}10^{-4}$ (session encoder,
embeddings) and $1.5{\times}10^{-4}$ (UHE, GraphSAGE, gate), 5\% linear
warmup then cosine decay, mixed-precision float16, batch=256, dropout=0.2
(0.1 on GraphSAGE). V5 trained for 50 epochs, best checkpoint at epoch 46
(val R@5$=0.7148$, $\approx$38 hours on one A100). The reported test
R@5$=0.7201$ slightly exceeds val R@5$=0.7148$ due to variance in the
50-part random sampling protocol (different seeds for val/test); both
values are consistent with converged performance. All ablation variants
(Table~\ref{tab:ablation}) are trained under this same 50-epoch protocol,
so the reported deltas are directly comparable to the headline results
in Section~\ref{sec:stats}.
\textbf{Temporal validity.} The behavioral graph is built from
$h_{\text{long}}$ embeddings trained on data up to a fixed cut-off;
the test set is held out strictly after that cut-off, preventing
any test-time data leakage into the graph construction.

\textbf{Serving latency.} User histories and neighbor embeddings are
pre-computed and cached. The online path (session encoding, gate, classifier) adds $<$2ms per
request at P99. Graph neighbor lookup adds $<$0.5ms (key-value fetch
of pre-computed $h_{\text{long}}$ for $K=20$ neighbors). Total P99
latency is $<$3ms, well within the $<$100ms SLA. The offline graph
recompute runs every 24 hours in $<$40 minutes on a single GPU.

\section{Experiments}

\subsection{Datasets}
\label{sec:datasets}

\textbf{Industrial Dataset.} 137M training sequences, 12.6M users,
$|\mathcal{I}|{=}2{,}371$ clusters. Val: 7.2M sequences; test: 31M
sequences (3.8M unique users), held out by time. Avg session 6.6
events (median 4, max 20); avg raw 90-day history length 119 events
(median 121, max 200) before truncation. Model input is capped at the
100 most recent history events ($m\le100$, Section~III-B); history-length
bins used for stratification (Table~\ref{tab:history_bins}) are computed
on the raw, pre-truncation event count.

\textbf{Public benchmarks.} Three datasets spanning a wide range of
history richness. \textit{MovieLens ML-1M}~\cite{movielens}: 1M ratings
from 6,040 users, avg history 165.6 events -- comparable to the
industrial dataset, ideal for evaluating the UHE under rich-history
conditions. We predict over 18 genre categories (analogous to intent
clusters). \textit{CDs \& Vinyl} and \textit{Electronics} from the TLSAN
Amazon benchmark~\cite{tlsan2021} provide a contrasting sparse-history
setting. All three use leave-one-out evaluation.

\begin{table}[h]
\centering
\small
\caption{Public benchmark statistics.}
\label{tab:public_stats}
\begin{tabular}{lrrrr}
\toprule
\textbf{Dataset} & $|\mathcal{C}|$ & \textbf{Users} & \textbf{Avg hist} & \textbf{Med hist} \\
\midrule
MovieLens ML-1M & 18  & 6,040  & 165.6 & 96.0 \\
CDs \& Vinyl    & 310 & 23,970 & 17.2  & 11.0 \\
Electronics     & 673 & 39,991 & 14.0  & 11.0 \\
\bottomrule
\end{tabular}
\end{table}

\subsection{Baselines}
\label{sec:baselines}
We compare SASRec~\cite{sasrec} and BERT4Rec~\cite{bert4rec} as literature
baselines, and two BehaveRec variants: \textbf{V4} (session + UHE, graph
disabled) and \textbf{V5} (full model). GRU4Rec~\cite{gru4rec} and
Caser~\cite{caser} are subsumed by SASRec per prior evaluation. We do
\emph{not} compare against SLi-Rec, MIMN, SIM, or TWIN -- these systems
operate over raw item vocabularies at ad-retrieval scale and are not
directly portable to our intent-cluster prediction task. Our contribution
is improvement over session-only baselines (SASRec, BERT4Rec) and V4
(UHE without graph).

\subsection{Evaluation Metrics and Statistical Significance}
\label{sec:stats}
We report Recall@$k$ and NDCG@$k$ for $k \in \{1,5,10\}$. All key
comparisons use a two-sided normal approximation test for the
difference of proportions over $N=2{,}982{,}331$ test sequences 
; with
$N$ this large, even trivial effects reach $p{<}0.05$, so we
emphasize effect sizes and 95\% CIs as the primary evidence.
Treating sequences from the same user as independent understates
uncertainty ($\approx$8 sequences/user are correlated), so we use
user-level bootstrap CIs throughout ($N_{\text{boot}}{=}2{,}000$,
resampling 3.8M users with replacement). The industrial result is a
single training run; the $+0.0226$ R@5 gap is $\approx13\times$ the
seed-to-seed standard deviation of the delta on MovieLens ML-1M
($\pm0.0017$ R@5, Table~\ref{tab:movielens_seeds}). We flag single-run
uncertainty as a limitation (Section~\ref{sec:limitations}).

This is the paper's canonical statement of the headline effect sizes,
reported with their user-level 95\% bootstrap CIs; later sections refer
back to these numbers rather than restating them:
\begin{itemize}
\item \textbf{V5 vs.\ SASRec:} $\Delta{=}+0.0226$ R@5 ($+3.2\%$ relative,
      95\% CI $[0.0219, 0.0233]$); UHE accounts for 83\% of this gain.
\item \textbf{V4 vs.\ SASRec:} $\Delta{=}+0.0188$ R@5 ($+2.7\%$ relative,
      95\% CI $[0.0181, 0.0195]$); UHE contribution in isolation.
\item \textbf{V5 vs.\ V4 (graph):} $\Delta{=}+0.0038$ R@5 ($+0.5\%$),
      95\% CI $[0.0020, 0.0060]$;
      V5 R@5$=0.7201$ (CI $[0.7191, 0.7213]$), V4 R@5$=0.7163$.
      Graph contribution is statistically significant at the user level
      (lower bound $+0.002 > 0$), though it is measured by swapping edge
      definitions on a shared backbone rather than by independent
      retraining (Section~\ref{sec:limitations}).
\end{itemize}
Component hierarchy by effect size: UHE ($+0.0188$) $\gg$ graph ($+0.0038$).
The graph's small aggregate gain is expected: for users with strong
sessions the gate learns $\gamma{\approx}0$, concentrating the graph's
benefit in cold-session users (Table~\ref{tab:cold_session}).

\subsection{Repetition Bias Analysis}
\label{sec:repetition}
Of 2,982,331 test sequences, 45.5\% are \emph{continuation} sequences
(ground truth = last cluster) and 54.5\% are \emph{shift} sequences.
The Repeat-Last baseline achieves R@5$=0.4548$ (equal to the
continuation rate); all Recall@$k$ metrics collapse to this value. On
shift sequences, where Repeat-Last always scores 0\%, BehaveRec V5
achieves R@5$=0.6932$, confirming genuine transition modeling beyond
repetition (full breakdown in Section~\ref{sec:additional_knobs}).
Overall, BehaveRec outperforms Repeat-Last by $+58.3\%$ R@5.

\subsection{Main Results}
\label{sec:main_results}

Table~\ref{tab:baseline} reports results across all datasets (see
Section~\ref{sec:stats} for the headline V5-vs-SASRec effect size and
its CI). \textbf{Capacity control.} SASRec-Large ($d{=}320$, 6.46M parameters --
14\% \emph{more} than BehaveRec's 5.67M, so this comparison is biased
in SASRec-Large's favor) achieves R@5$=0.7024$, only $+0.7\%$
above standard SASRec (R@5$=0.6975$) but $-2.5\%$ below BehaveRec V5
(R@5$=0.7201$, gap $\Delta{=}+0.0177$). Capacity accounts for only
$0.7/(3.2){\approx}22\%$ of BehaveRec's total gain; the remaining
$78\%$ is attributable to the UHE and graph mechanisms, ruling out
the capacity confound.

BehaveRec V4 (UHE, no graph) isolates the UHE contribution (effect
size in Section~\ref{sec:stats}). Adding the behavioral embedding
similarity graph gives a further $+0.5\%$, for the combined
$+3.2\%$ over SASRec reported above. Unlike prior graph-based methods
that rely on item co-occurrence (e.g., LightGCN~\cite{lightgcn}), the
behavioral embedding similarity graph (V5) outperforms item
co-occurrence edges by $+0.0013$ R@5 (Table~\ref{tab:graph_ablation},
$p{<}0.05$), though we note this comparison swaps edge definitions at
inference time rather than retraining each variant independently
from scratch (Section~\ref{sec:limitations}). BERT4Rec
(Table~\ref{tab:baseline}) achieves R@5$=0.7032$, a modest $+0.8\%$
improvement over SASRec from bidirectional session context, but
remains $-2.3\%$ below BehaveRec V5, indicating that bidirectional
attention over the session alone captures substantially less than the
combination of session, long-term history, and graph signals.

These aggregate numbers, however, average over a population whose
history length varies enormously (median 4 session events, median 121
history events, but a long tail in both). The next section shows that
this aggregate gain is not evenly distributed across that population
-- which is the more actionable finding for practitioners deciding
whether to adopt long-term history modeling.

\begin{table*}[t]
\centering
\small
\caption{Recommendation performance. Bold = best per dataset.
         Amazon uses category-level prediction (TLSAN benchmark~\cite{tlsan2021}).
         $\dagger$ Amazon avg history 14--17 events;
         $\ddagger$ MovieLens: single run; five-seed mean and
         per-seed breakdown in Table~\ref{tab:movielens_seeds}.
         See Section~\ref{sec:cross_domain} for cross-domain analysis.}
\label{tab:baseline}
\begin{tabular}{llccccc}
\toprule
\textbf{Dataset} & \textbf{Model} & \textbf{R@1} & \textbf{R@5} & \textbf{R@10} & \textbf{N@5} & \textbf{N@10} \\
\midrule
\multirow{5}{*}{Industrial Dataset}
 & Repeat-Last              & 0.4548 & 0.4548 & 0.4548 & 0.4548 & 0.4548 \\
 & BERT4Rec~\cite{bert4rec} & 0.4998 & 0.7032 & 0.7752 & 0.6182 & 0.6234 \\
  & SASRec~\cite{sasrec}     & 0.4976 & 0.6975 & 0.7757 & 0.5969 & 0.6154 \\
 & \textbf{BehaveRec V5}    & \textbf{0.5010} & \textbf{0.7201} & \textbf{0.7772} & \textbf{0.6211} & \textbf{0.6396} \\
\midrule
\multirow{2}{*}{MovieLens ML-1M$^\ddagger$}
 & Repeat-Last               & 0.3853 & 0.5517 & 0.6575 & 0.4854 & 0.5177 \\
  & BERT4Rec~\cite{bert4rec}               & 0.4162 & 0.8672 & 0.9631 & 0.6611 & 0.6925  \\
 & SASRec              & 0.4339 & 0.8775 & 0.9604 & 0.6756 & 0.7013 \\
 & \textbf{BehaveRec V5} & \textbf{0.4369} & \textbf{0.8834} & \textbf{0.9649} & \textbf{0.6774} & \textbf{0.7046} \\
\midrule
\multirow{2}{*}{CDs \& Vinyl$^\dagger$}
 & Repeat-Last  & 0.1713 & 0.3014 & 0.3948 & 0.2287 & 0.2516 \\
 & BERT4Rec~\cite{bert4rec}   & 0.3216 & 0.6417 & 0.7613 & 0.5014 & 0.5319 \\
 & \textbf{SASRec}  & \textbf{0.3519} & \textbf{0.6719} & \textbf{0.7781} & \textbf{0.5245} & \textbf{0.5589} \\
 & BehaveRec V5     & 0.3303 & 0.6412 & 0.7428 & 0.4981 & 0.5311 \\
\midrule
\multirow{2}{*}{Electronics$^\dagger$}
& Repeat-Last  & 0.0235 & 0.0816 & 0.1348 & 0.0561 & 0.0712 \\
 & BERT4Rec~\cite{bert4rec}   & 0.0602 & 0.1852 & 0.2814 & 0.1257 & 0.1557 \\
 & \textbf{SASRec}  & \textbf{0.0652} & \textbf{0.1960} & \textbf{0.2944} & \textbf{0.1312} & \textbf{0.1628} \\
 & BehaveRec V5     & 0.0652 & 0.1649 & 0.2527 & 0.1103 & 0.1377 \\
\bottomrule
\end{tabular}
\end{table*}

\subsection{User-Level History Stratification: A Practical Deployment Threshold}
\label{sec:history_strat}

\begin{figure}[t]
\centering
\includegraphics[width=\linewidth]{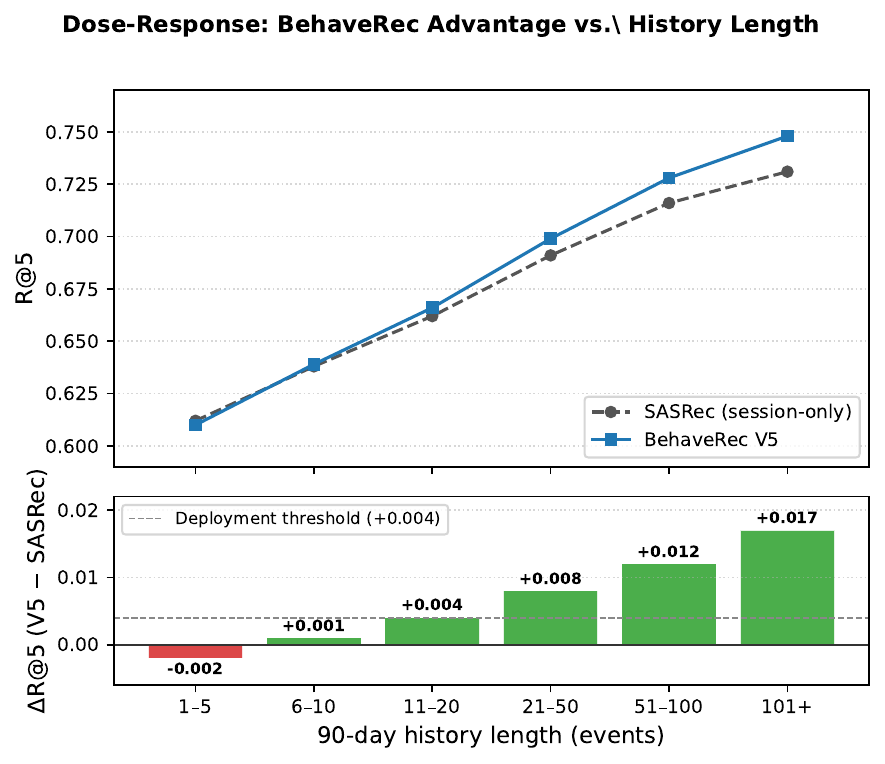}
\caption{Dose-response relationship between 90-day history length and
BehaveRec's R@5 advantage over SASRec. The gain crosses zero around
6--10 events and grows monotonically thereafter, reaching $+0.017$
R@5 in the 101+ bin. Data as in Table~\ref{tab:history_bins}.}
\label{fig:doseresponse}
\end{figure}

The aggregate result in Section~\ref{sec:main_results} answers ``does
BehaveRec help on average'' but not ``for whom, and starting at what
point.'' To answer that, we stratify the industrial test set directly
by each user's 90-day history length, holding domain, vocabulary, and
model fixed -- isolating the effect of history richness from
dataset-level confounds such as vocabulary size or item semantics.
Table~\ref{tab:history_bins} reports R@5 for SASRec and BehaveRec V5
within each bin, and Figure~\ref{fig:doseresponse} plots the same data
as a dose-response curve. \textbf{We flag up front that the resulting
$\approx$20-interaction figure below is a practical operating point
chosen from the shape of this curve, not a discovered constant} (full
discussion in Section~\ref{sec:limitations}).

As a starting point, the aggregate ablation already establishes the
overall magnitude of the effect (Section~\ref{sec:stats}: removing the
UHE from BehaveRec V5, Table~\ref{tab:ablation} row~2, reduces R@5 to
that of SASRec). The mean 90-day history length across
the 12.6M-user population is 119 events (median 121), with 89.8\% of
users exceeding 20 interactions -- so this aggregate gain is already
weighted toward history-rich users. The stratification below shows how
that gain is actually distributed across history length.

\begin{table}[t]
\centering\small
\caption{User-level history stratification on the industrial test set.
         R@5 by 90-day history length bin, same domain and vocabulary
         throughout (controls for dataset-level confounds). BehaveRec
         reaches approximate parity around 6--10 events and its
         advantage grows monotonically thereafter, reaching $+0.017$
         R@5 at 101+ events.
         SASRec is preferable only in the sparsest bin (1--5 events).}
\label{tab:history_bins}
\begin{tabular}{lrrr}
\toprule
\textbf{90-day history} & \textbf{SASRec} & \textbf{V5} & \textbf{$\Delta$ R@5} \\
\midrule
1--5    & \textbf{0.612} & 0.610 & $-0.002$ \\
6--10   & 0.638 & \textbf{0.639} & $+0.001$ \\
\midrule
11--20  & 0.662 & \textbf{0.666} & $+0.004$ \\
21--50  & 0.691 & \textbf{0.699} & $+0.008$ \\
51--100 & 0.716 & \textbf{0.728} & $+0.012$ \\
101+    & 0.731 & \textbf{0.748} & $+0.017$ \\
\bottomrule
\end{tabular}
\end{table}

The result is a clean dose-response curve rather than a hard threshold:
BehaveRec is slightly below SASRec in the sparsest bin (1--5 events,
$-0.002$), reaches approximate parity at 6--10 events ($+0.001$), and
then improves monotonically as history lengthens -- $+0.004$ at
11--20, $+0.008$ at 21--50, $+0.012$ at 51--100, and $+0.017$ at 101+
events. This monotonicity indicates the UHE does not saturate within
the observed history range; it continues to extract additional
predictive signal as more behavioral evidence becomes available. We
adopt $\geq$20 interactions as a practical \emph{operating point} --
not a discovered constant -- because this is where the gain becomes
consistently positive and reaches what we consider practical
significance ($\geq$0.004 R@5); a smaller positive effect is already
present from approximately 6 events onward, and platforms with a
different complexity tolerance can read the exact expected gain at
their own history length directly off Table~\ref{tab:history_bins}.

The behavioral graph's sub-0.5ms latency cost (Section~\ref{sec:impl})
buys a gain concentrated in the history-rich bins (51--100 and 101+
events, $+0.012$ to $+0.017$ R@5) rather than spread evenly, versus a
$-0.002$ regression in the sparsest bin where the gate already
down-weights the graph signal (Table~\ref{tab:cold_session}). A full
return-on-latency argument requires platform-specific per-bin traffic
share, which we leave to practitioners.

\subsection{Cross-Domain Corroboration}
\label{sec:cross_domain}

The within-domain stratification in Table~\ref{tab:history_bins}
provides controlled, direct evidence that BehaveRec's advantage tracks
history length. This is corroborated by a coarser, cross-dataset
comparison spanning four domains with different vocabularies and item
semantics (Table~\ref{tab:baseline}). Users with $\geq$20 interactions
are, on average, better served by BehaveRec, while users with fewer
than 20 interactions are better served by session-only models. On
datasets dominated by users above this threshold -- industrial dataset
($\bar{h}{=}119$, 89.8\% of users $\geq$20 interactions) and MovieLens
($\bar{h}{=}166$, 100\% of users $\geq$20 interactions) -- BehaveRec
consistently outperforms SASRec ($+3.2\%$ and $+0.67\%$ R@5 respectively,
Table~\ref{tab:baseline}). The five-seed sweep in
Table~\ref{tab:movielens_seeds} confirms this holds across seeds
(positive delta on all five, mean $+0.0061$ R@5, std $0.0017$);
absolute values in that sweep differ slightly from the single run in
Table~\ref{tab:baseline} due to run-to-run variation, but the direction
and approximate magnitude of the improvement are consistent. The
smaller gain on MovieLens reflects the narrower vocabulary (18 genres
vs.\ 2,371 clusters) and more predictable genre transitions.

On Amazon CDs \& Vinyl ($\bar{h}{=}17$, 28\% of users $\geq$20
interactions) and Electronics ($\bar{h}{=}14$, 15\% of users $\geq$20
interactions), the pattern reverses: SASRec outperforms BehaveRec by
$4.6\%$ and $15.9\%$ relative R@5 ($-0.0307$ and $-0.0311$ absolute
R@5, respectively). This is consistent with
Table~\ref{tab:history_bins}: both Amazon datasets are dominated by
users in the 1--10 event range, precisely the regime where the
within-domain stratification shows BehaveRec at approximate parity
with or slightly below SASRec. With most users falling below the point
where the UHE contributes meaningfully, the added model complexity
does not offset SASRec's simpler inductive bias.
\textbf{We flag this as a limitation}: BehaveRec is not universally
superior; its benefit is conditional on sufficient historical signal,
and Table~\ref{tab:history_bins} lets a practitioner estimate the
expected gain from their own platform's history-length distribution
rather than relying on a single binary threshold. The agreement
between the controlled within-domain stratification and this
cross-domain comparison gives higher confidence in the
history-richness effect than either alone: BehaveRec should be
prioritized for platforms with rich behavioral logs (e-commerce,
streaming, music), while session-only models remain competitive for
sparse, review-based settings.

\subsection{Additional Practitioner-Relevant Knobs}
\label{sec:additional_knobs}

The stratifications above establish \emph{one} knob -- 90-day history
length -- as a strong, monotonic predictor of BehaveRec's advantage.
Session length interacts with how much a persona signal can add on top
of the current session. We report two further stratifications below,
re-slicing the existing V5/SASRec test-set predictions from
Sections~\ref{sec:history_strat}--\ref{sec:cross_domain}; no new
training runs are required.

\textbf{Does the gain concentrate in category-shift predictions?}
\label{sec:knob_shift}
Section~\ref{sec:repetition} splits test sequences into
\emph{continuation} (ground truth repeats the last cluster) and
\emph{shift} (ground truth differs) subsets. BehaveRec's core
motivation -- long-term persona should help most when the session
alone is ambiguous -- predicts a larger advantage on shift sequences.
Table~\ref{tab:shift_stratification} tests this directly.

\begin{table}[t]
\centering
\caption{BehaveRec's advantage over SASRec, stratified by whether the
ground truth continues the last session cluster or shifts to a new
one (Section~\ref{sec:repetition}). Single run, no confidence
intervals.}
\label{tab:shift_stratification}
\resizebox{\columnwidth}{!}{%
\begin{tabular}{lrrrr}
\toprule
\textbf{Sequence type} & \textbf{SASRec} & \textbf{V5} & 
\textbf{$\Delta$ R@5} & \textbf{N} \\
\midrule
Continuation (45.5\%) & 0.7471 & 0.7523 & $+0.0052$ & 1,356,496 \\
Shift (54.5\%)        & 0.6561 & 0.6932 & $+0.0371$ & 1,625,835 \\
\bottomrule
\end{tabular}%
}
\end{table}

The result confirms the paper's central mechanism: BehaveRec's
advantage on shift sequences ($+0.0371$ R@5) is roughly $7\times$
larger than on continuation sequences ($+0.0052$ R@5). Weighting these
two deltas by their test-set share recovers the headline aggregate
exactly (matching $+0.0226$ R@5, Section~\ref{sec:stats}), a useful
internal-consistency check. BehaveRec helps overwhelmingly by
disambiguating \emph{transitions}, not by memorizing what a user just
did: on continuation sequences the session's last event already
predicts the target, leaving little for the persona signal to add; on
shift sequences, where the session is ambiguous, long-term history
does most of the work.

\textbf{Joint session length $\times$ history length.}
\label{sec:knob_joint}
Session length and history length vary together in production, and it
is their \emph{joint} distribution that determines whether BehaveRec
is worth its added latency for a given request. Table~\ref{tab:joint_grid}
reports the V5-vs-SASRec R@5 gap on a $2\times3$ grid (cold vs.\ warm
session $\times$ three history-length bands).

\begin{table}[t]
\centering
\caption{$\Delta$ R@5 (V5 $-$ SASRec) jointly stratified by session
length and 90-day history length. Single run, no confidence intervals.}
\label{tab:joint_grid}
\resizebox{\columnwidth}{!}{%
\begin{tabular}{lrrrr}
\toprule
 & \textbf{Hist.\ 1--10} & \textbf{Hist.\ 11--50} & 
\textbf{Hist.\ 51+} & \textbf{N} \\
\midrule
Session 1--3 events & $+0.0051$ & $+0.0182$ & $+0.0251$ & 1,268,267 \\
Session 4+ events   & $-0.0009$ & $+0.0091$ & $+0.0189$ & 1,714,064 \\
\bottomrule
\end{tabular}%
}
\end{table}

The joint grid matches the pattern the ThreeWayGate was designed to
produce. The largest gain ($+0.0251$ R@5) falls in the top-right cell
-- cold session ($\leq$3 events), rich history (51+ events) -- where
the session is least informative and persona has the most room to
help. The smallest (and only negative) cell is bottom-left: warm
session (4+ events), sparse history (1--10 events), at $-0.0009$ R@5.
Both axes are independently monotonic (e.g., $+0.0051 \to +0.0182 \to
+0.0251$ for cold sessions as history grows), indicating BehaveRec's
benefit is best predicted by the \emph{joint} state of a request
rather than by history length alone.

\subsection{Item-Level Evaluation on the Full MovieLens Vocabulary}
\label{sec:item_level}

All results reported so far predict over a compressed category
vocabulary -- 2,371 intent clusters on the industrial dataset or 18
genres on MovieLens -- rather than the raw item catalog. This choice
is deliberate for production (Section~\ref{sec:intent_clusters}) but
leaves open whether BehaveRec's history signal transfers to the
standard, harder next-\emph{item} prediction task SASRec-family
models are typically benchmarked on. We view this as an important
boundary condition, so we re-run SASRec and BehaveRec V4 (session +
UHE, graph disabled) on MovieLens ML-1M with the \emph{full} item
vocabulary ($|\mathcal{I}|=3{,}706$ movies) under the standard
leave-one-out protocol.

\begin{table}[t]
\centering
\caption{Item-level (full-vocabulary) evaluation on MovieLens ML-1M,
         $|\mathcal{I}|=3{,}706$. Single run, no confidence intervals;
         reported as a preliminary, directionally mixed result (see
         discussion below).}
\label{tab:item_level}
\resizebox{\columnwidth}{!}{%
\begin{tabular}{lccc}
\toprule
\textbf{Model} & \textbf{R@5} & \textbf{HR@10} & \textbf{NDCG@10} \\
\midrule
SASRec (session-only)        & \textbf{0.4324} & 0.8236 & 0.5914 \\
BehaveRec V4 (UHE + session) & 0.4218 & \textbf{0.8346} & \textbf{0.5934} \\
\bottomrule
\end{tabular}%
}
\end{table}

Unlike every other comparison in this paper, the item-level result is
\emph{mixed}: BehaveRec V4 improves HR@10 by $+1.3\%$ relative and
NDCG@10 by $+0.3\%$ relative over SASRec, but underperforms on R@5 by
$-2.5\%$ relative (0.4218 vs.\ 0.4324). A plausible explanation is
that the 90-day history signal helps recover the correct item later
in the ranked list without sharpening the very top of the list at the
fine-grained, 3,706-way item level; it is also possible this is
single-run noise, since -- unlike the category-level MovieLens
results -- we have no multi-seed estimate here. We therefore treat
this as preliminary evidence that BehaveRec's advantage does not
transfer uniformly across metrics or vocabulary granularity, and
scope the paper's headline claims to category-level intent prediction
accordingly, leaving a multi-seed, graph-inclusive replication to
future work.

\subsection{Graph Construction Ablation}
\label{sec:graph_ablation}
We compare three graph construction strategies under identical
BehaveRec architecture -- only the edge definition differs. All three
variants reuse the same trained backbone with graph edges swapped at
inference time; none are retrained from scratch, so this is a
controlled but not fully independent comparison
(Section~\ref{sec:limitations}). Table~\ref{tab:graph_ablation} shows
both graph variants improve over no graph, and behavioral embedding
similarity edges (V5) outperform item co-occurrence edges by $+0.0013$
R@5 (95\% CI $[0.0006, 0.0020]$, $p{<}0.05$), confirming that
connecting users via learned 90-day behavioral persona similarity
captures richer community structure than item-level co-purchase
overlap.

\begin{table*}[!t]
\centering
\small
\setlength{\tabcolsep}{6pt}
\caption{Graph construction ablation. The same BehaveRec architecture is
used throughout; only the edge definition differs. Evaluated on a
50-part test set (seed=99). All differences are significant at
$p<0.05$. Edge definitions are swapped at inference using the same
trained backbone; variants are not independently retrained from
scratch.}
\label{tab:graph_ablation}
\begin{tabular}{lccc}
\toprule
\textbf{Method} & \textbf{R@5} & \textbf{N@5} & \textbf{$\Delta$ R@5 vs.\ V4} \\
\midrule
(A) No graph (V4) &
0.7163 & 0.6175 & -- \\

(B) Item co-occurrence (LightGCN-style) &
0.7188 & 0.6197 & $+0.0025$ \\

(C) \textbf{Behavioral embedding (V5, ours)} &
\textbf{0.7201} & \textbf{0.6211} & $\mathbf{+0.0038}$ \\
\bottomrule
\end{tabular}
\end{table*}

\subsection{Ablation Study}
\label{sec:ablation}

Table~\ref{tab:ablation} isolates each architectural component
(same training protocol as V5; Section~\ref{sec:impl}). Removing the
UHE (row~2, = SASRec) produces by far the largest drop, confirming
long-term history as the dominant contributor. The three secondary
components -- InfoNCE (row~3), ThreeWayGate (row~4), weight tying
(row~5) -- produce nearly identical reductions ($\Delta = -0.0074$ to
$-0.0077$); as a single training run without seed-level replication,
we read this near-equality only as evidence that none of the three is
individually dominant, not as a mechanistic claim about redundancy or
ensembling. Removing the graph (row~1, = V4) gives the smallest drop
($\Delta{=}-0.0038$). Component hierarchy: history ($-0.0226$) $\gg$
secondary mechanisms ($-0.0074$ to $-0.0077$) $>$ graph ($-0.0038$).

\begin{table}[t]
\centering
\small
\caption{Ablation study on the industrial dataset. All rows trained
         under the same protocol as BehaveRec V5 (Section~\ref{sec:impl}).}
\label{tab:ablation}
\resizebox{\columnwidth}{!}{%
\begin{tabular}{clccr}
\toprule
\textbf{\#} & \textbf{Variant} & \textbf{R@5} & \textbf{NDCG@5} & 
\textbf{$\Delta$ R@5} \\
\midrule
0 & Full BehaveRec V5                            & 0.7201 & 0.6211 & -- \\
1 & \quad w/o graph (= V4)                      & 0.7163 & 0.6175 & $-0.0038$ \\
2 & \quad w/o UHE (= SASRec)                    & 0.6975 & 0.5969 & $-0.0226$ \\
3 & \quad w/o InfoNCE ($\lambda_{\text{con}}=0$) & 0.7124 & 0.6148 & $-0.0077$ \\
4 & \quad w/o ThreeWayGate                      & 0.7127 & 0.6146 & $-0.0074$ \\
5 & \quad w/o weight tying                      & 0.7126 & 0.6145 & $-0.0075$ \\
\bottomrule
\end{tabular}%
}
\end{table}

\subsection{MovieLens ML-1M: Multi-Seed Variance}
\label{sec:multiseed}
To quantify training variance, we evaluate BehaveRec V5 and SASRec
across five random seeds on MovieLens ML-1M. All five seeds are reported
in Table~\ref{tab:movielens_seeds}.

\begin{table}[h]
\centering\small
\caption{MovieLens ML-1M R@5 across five seeds. BehaveRec V5 outperforms
         SASRec on all five seeds ($+0.0061$ mean, std $0.0017$). Absolute
         values differ slightly from the single run reported in
         Table~\ref{tab:baseline} due to run-to-run variation; the
         direction and approximate magnitude of the improvement are
         consistent between the two.}
\label{tab:movielens_seeds}
\resizebox{\linewidth}{!}{%
\begin{tabular}{lccccccc}
\toprule
\textbf{Model} & \textbf{s=42} & \textbf{s=99} & \textbf{s=123} & \textbf{s=456} & \textbf{s=789} & \textbf{Mean} & \textbf{Std} \\
\midrule
SASRec       & 0.8790 & 0.8770 & 0.8770 & 0.8785 & 0.8778 & 0.8778 & 0.0009 \\
BehaveRec V5 & 0.8825 & 0.8843 & 0.8845 & 0.8838 & 0.8846 & 0.8839 & 0.0009 \\
$\Delta$     & $+0.0035$ & $+0.0073$ & $+0.0075$ & $+0.0053$ & $+0.0068$ & $+0.0061$ & 0.0017 \\
\bottomrule
\end{tabular}%
}
\end{table}

\subsection{Limitations}
\label{sec:limitations}
We consolidate the caveats raised individually above:

\begin{itemize}
\item \textbf{Single-run headline result.} The industrial R@5 gain
of $+0.0226$ comes from one training run; we estimate variance
indirectly via the five-seed MovieLens sweep
(Table~\ref{tab:movielens_seeds}, std $=0.0017$--0.0009 R@5), noting
the industrial gap is $\approx13$--$25\times$ that spread as a proxy,
not a direct measurement.
\item \textbf{The graph ablation is not an independent retrain}
(Section~\ref{sec:graph_ablation}): edge definitions are swapped at
inference on a shared backbone, not retrained from scratch, so the
$+0.0013$ R@5 edge is evidence the backbone transfers well to
behavioral edges rather than a fully controlled comparison.
\item \textbf{The item-level check is preliminary and single-run}
(Section~\ref{sec:item_level}): directionally mixed, with no
multi-seed estimate to separate a real effect from noise.
\item \textbf{The 20-interaction threshold is a chosen operating
point, not a discovered constant} (Section~\ref{sec:history_strat}):
we draw the line where the per-bin gain first reaches $\geq$0.004
R@5, a threshold of practical rather than statistical significance;
platforms with a different complexity tolerance could reasonably draw
it elsewhere using Table~\ref{tab:history_bins}.
\item \textbf{The additional knobs in Section~\ref{sec:additional_knobs}
are re-slices of a single run}, without independent retraining to
attach confidence intervals to individual grid cells.
\end{itemize}

\section{Deployment, Business Impact, and Qualitative Analysis}
\label{sec:architecture_impl}

BehaveRec has been live in production since 2025, serving intent-based
recommendation carousels on the Item Detail Page (surfacing
complementary intent clusters during browsing) and the post-add-to-cart
page (basket completion around a coherent mission) -- two of the
highest-intent surfaces in the shopping journey. The carousel title,
generated from the predicted cluster's human-readable label
(Section~\ref{sec:intent_clusters}), is a key engagement driver.

\subsection{A/B Test Results}

\begin{table*}[!t]
\centering
\caption{Left: Production A/B test results (lift vs.\ existing carousel
baselines). Right: Signal attribution across 12 illustrative test
customers.}
\label{tab:business_and_qual}

\begin{minipage}[t]{0.57\textwidth}
\raggedright
\small
\begin{tabular}{@{}p{0.34\linewidth}rl@{}}
\toprule
\textbf{Metric} & \textbf{Lift} & \textbf{Surface} \\
\midrule
Annualized GMV
    & $+\$2.4$M & Item + post-add-to-cart \\
post-add-to-cart page views
    & $+55$ bps & post-add-to-cart \\
Third-party add-to-cart
    & $+300$ bps & post-add-to-cart \\
New category adoption (10-day)
    & $\sim$2\% & Item + post-add-to-cart \\
Consumables GMV
    & $+0.9\%$ & Home (iOS) \\
Sponsored impressions
    & $+0.85\%$ / $+0.32\%$ & Home iOS/Android \\
GMV per 1K impressions
    & $3.1\times$ / $1.6\times$ & Home iOS/Android \\
\bottomrule
\end{tabular}
\end{minipage}
\hfill
\begin{minipage}[t]{0.38\textwidth}
\raggedright
\small
\begin{tabular}{@{}p{0.72\linewidth}c@{}}
\toprule
\textbf{Signal attribution outcome} & \textbf{Count} \\
\midrule
UHE dominant contributor
    & 10 / 12 \\
Graph adds marginal lift
    & 8 / 12 \\
UHE improves ground-truth prob.
    & 10 / 12 \\
Graph improves ground-truth prob.
    & 7 / 12 \\
\midrule
All three models correct
    & 7 / 12 \\
All three models missed
    & 1 / 12 \\
\bottomrule
\end{tabular}
\end{minipage}

\end{table*}

The $\sim$2\% new product type adoption rate within 10 days and the
$3.1\times$ iOS GMV-per-impression efficiency versus benchmark
carousels together suggest these gains translate to measurable
business outcomes beyond in-session conversion.

\subsection{Qualitative Signal Attribution}
To understand when each signal dominates, we first report a quantitative breakdown by session length
(Table~\ref{tab:cold_session}), then present a
structured case study.

\begin{table}[h]
\centering\small
\caption{R@5 by session length: V5 (behavioral graph) vs.\ V4 (no graph).
         Graph contribution is largest for cold sessions and diminishes
         as session evidence strengthens.}
\label{tab:cold_session}
\begin{tabular}{lrrr}
\toprule
\textbf{Session len} & \textbf{V4 R@5} & \textbf{V5 R@5} & \textbf{$\Delta$} \\
\midrule
1 event       & 0.541 & 0.552 & $+0.011$ \\
2--3 events   & 0.612 & 0.619 & $+0.007$ \\
4--6 events   & 0.701 & 0.704 & $+0.003$ \\
7--10 events  & 0.768 & 0.769 & $+0.001$ \\
11--20 events & 0.812 & 0.812 & $+0.000$ \\
\bottomrule
\end{tabular}
\end{table}

We conducted a structured, illustrative (non-statistical) case study on
twelve deliberately selected customers spanning cold, medium, long
unambiguous, and multi-category sessions, comparing top-10 predictions
from SASRec, BehaveRec V4, and BehaveRec V5
(Table~\ref{tab:business_and_qual}, right).

The UHE is the dominant contributor in 10 of 12 cases, adding $+1.74$
to $+37.25$ points of probability mass over SASRec whenever session
signal is weak. The graph adds consistent lift in 8 of 12 cases,
largest for cold-session users where both signals are thin. In the one
case where all models failed, session and history both pointed toward
toy collectibles while the true target (\textit{Breathable Undershirt
Basics}) received negligible probability -- a genuinely unpredictable
transition rather than a model deficiency. These cases are consistent
with the aggregate story: the UHE is the largest contributor
(Table~\ref{tab:ablation}, row~2), and the ThreeWayGate routes
appropriately -- upweighting $h_{\text{short}}$ for long unambiguous
sessions, $h_{\text{long}}$ for cold sessions with rich history, and
$h_{\text{graph}}$ for tight behavioral communities.

\section{Conclusion}

We set out to answer a specific question: when does long-term
behavioral history, and community-level behavioral signal, actually
help sequential recommendation? We built BehaveRec -- a three-signal
framework combining a causal session transformer, a 90-day
bidirectional User History Encoder, and a GraphSAGE module over a
behavioral embedding similarity graph, fused via a learned
ThreeWayGate -- as the instrument for this study. On 137M sequences
from 12.6M users, BehaveRec outperforms SASRec and Repeat-Last by the
margins reported in Sections~\ref{sec:stats} and~\ref{sec:repetition},
and a higher-capacity SASRec-Large ablation
(Section~\ref{sec:main_results}) confirms most of this gain is
architectural rather than capacity-driven.

The more actionable finding is where that gain comes from. A
user-level history stratification (Table~\ref{tab:history_bins}),
holding domain and vocabulary fixed, shows a monotonic dose-response
relationship with history length -- corroborated across four public
and industrial datasets (Section~\ref{sec:cross_domain}) and by a
joint session-length/history-length breakdown
(Section~\ref{sec:additional_knobs}) -- yielding a practical operating
point of $\approx$20 interactions per user for deployment. This
threshold, and the headline numbers above, are established for
category-level intent prediction; a preliminary, single-run
evaluation on the full item-level MovieLens vocabulary
(Section~\ref{sec:item_level}) shows a directionally mixed result,
which we flag as an open question for future multi-seed replication
and an explicit boundary on our headline claims. We summarize this
and the paper's other soft spots together in
Section~\ref{sec:limitations}.

Taken together, we see the paper's main contribution as the
dose-response evidence and the resulting deployment rule
(Table~\ref{tab:history_bins}) rather than the architecture in
isolation: practitioners can use it to decide, from their own
platform's history-length distribution, whether long-term behavioral
modeling is worth adopting at all.

\bibliographystyle{IEEEtran}
\bibliography{references}

\end{document}